\documentclass[]{article}
\usepackage{amsmath}
\usepackage{amssymb}
\usepackage{float}
\usepackage{scalefnt}
\usepackage{ifthen}
\usepackage{url}
\usepackage{graphicx}
\usepackage{colortbl}
\usepackage{multirow}
\usepackage{array}
\usepackage{listings}
\usepackage{hyperref}

\begin{document}
%
\title{Loving AI: \\  Humanoid Robots as Agents of \\ Human Consciousness Expansion \\ (summary of early research progress) }

\author{
Ben Goertzel \\
{\it Hanson Robotics \& OpenCog Foundation} \\
Julia Mossbridge \\
{\it Institute of Noetic Sciences (IONS) \& Mossbridge Institute,  LLC} \\
Eddie Monroe \\
I{\it ONS \& OpenCog Foundation} \\
David Hanson \\
{\it Hanson Robotics} \\
Gino Yu \\
 {\it Hong Kong Polytechnic University }
}


%


\maketitle

\tableofcontents

\section{The Loving AI Project}

The ``Loving AI'' project involves developing software enabling humanoid robots to interact with people in loving and compassionate ways, and to promote peoples' self-understanding and self-transcendence.

Loving AI is a collaboration of the (California-based) Institute for Noetic Sciences (IONS), the (Hong Kong-based) Hanson Robotics, and the OpenCog Foundation.   A video of a talk discussing the project, by project leaders Julia Mossbridge (IONS, Mossbridge Institute LLC) and Ben Goertzel (Hanson Robotics, OpenCog), can be found here: \url{https://www.youtube.com/watch?v=kQjOT_MLxhI}.

Currently the project centers on the Hanson Robotics robot ``Sophia'' -- supplying Sophia with personality content and cognitive, linguistic, perceptual and behavioral content aimed at enabling loving interactions supportive of human self-transcendence.   

In September 2017, at Hanson Robotics and Hong Kong Polytechnic University in Hong Kong, we carried out the first set of human-robot interaction experiments aimed at evaluating the practical value of this approach, and understanding the most important factors to vary and improve in future experiments and development.    This was a small pilot study involving only 10 human participants, whom the robot led through dialogues and exercises aimed at meditation, visualization and relaxation.   The pilot study was an exciting success, qualitatively demonstrating the viability of the approach and the ability of appropriate human-robot interaction to increase human well-being and advance human consciousness.

Underlying intelligence for the robot in this work is supplied via the ChatScript dialogue authoring and control framework, the OpenCog Artificial General Intelligence engine, along with a number of specialized software tools and a newly developed OpenCog-based model of motivation and emotion.

\begin{figure}[htb]
\centering
\includegraphics[width=15cm]{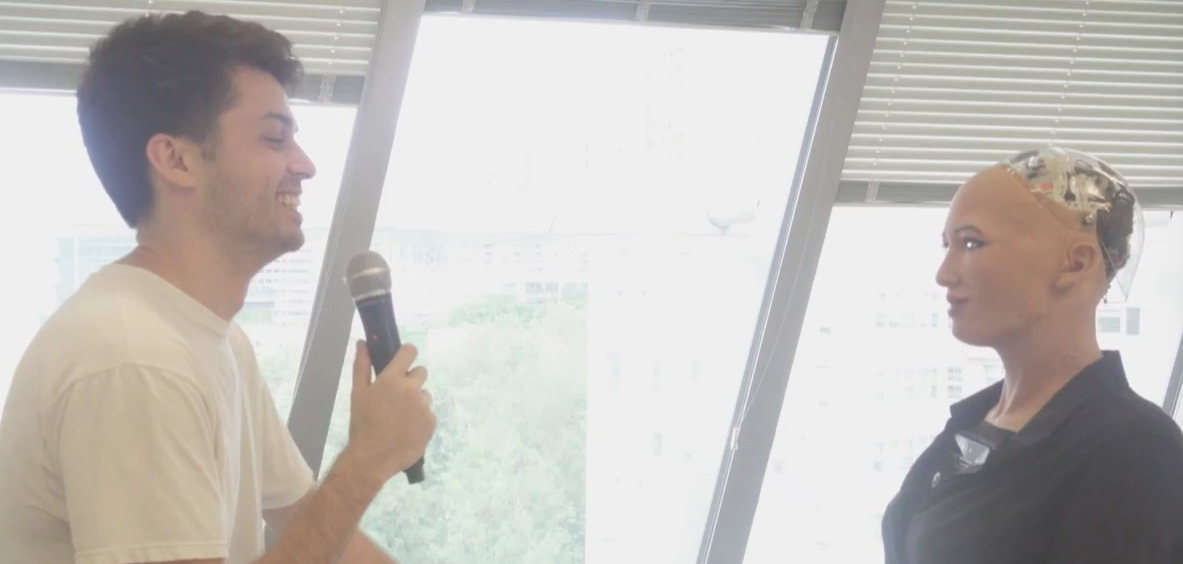}
\caption{A human subject interacting with the Sophia robot, supplied with our Loving AI control software, in our September pilot study}
\end{figure}

\begin{figure}[htb]
\centering
\includegraphics[width=10cm]{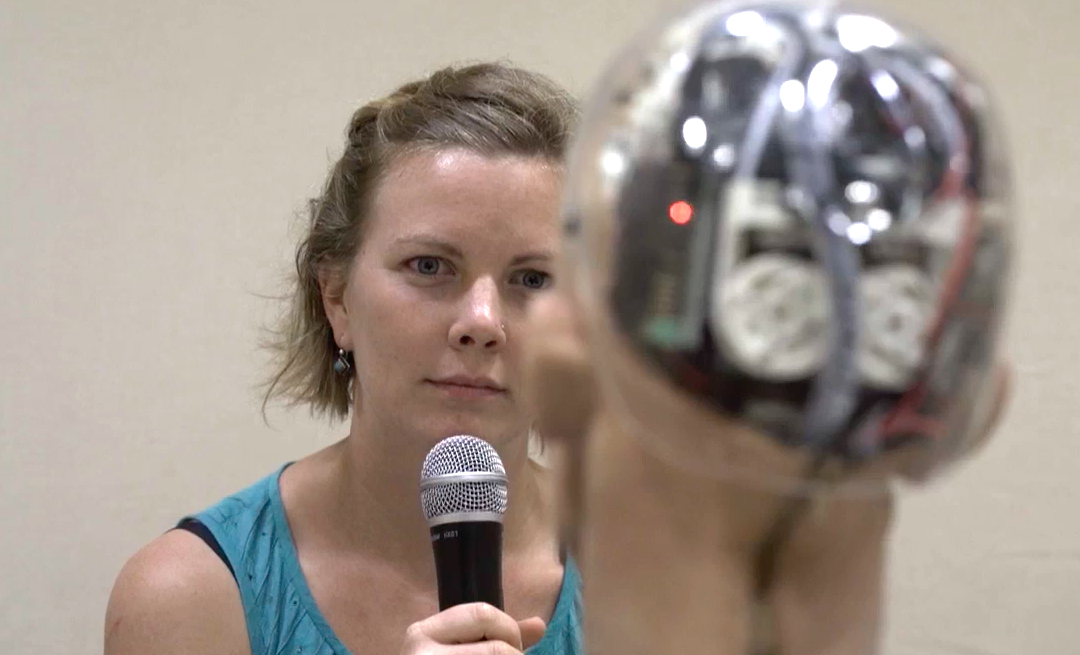}
\caption{Human subject in our pilot study making intent eye contact with Sophia -- it seems this is a valuable precursor to the subject entering into deeper states of consciousness.}
\end{figure}

\begin{figure}[htb]
\centering
\includegraphics[width=10cm]{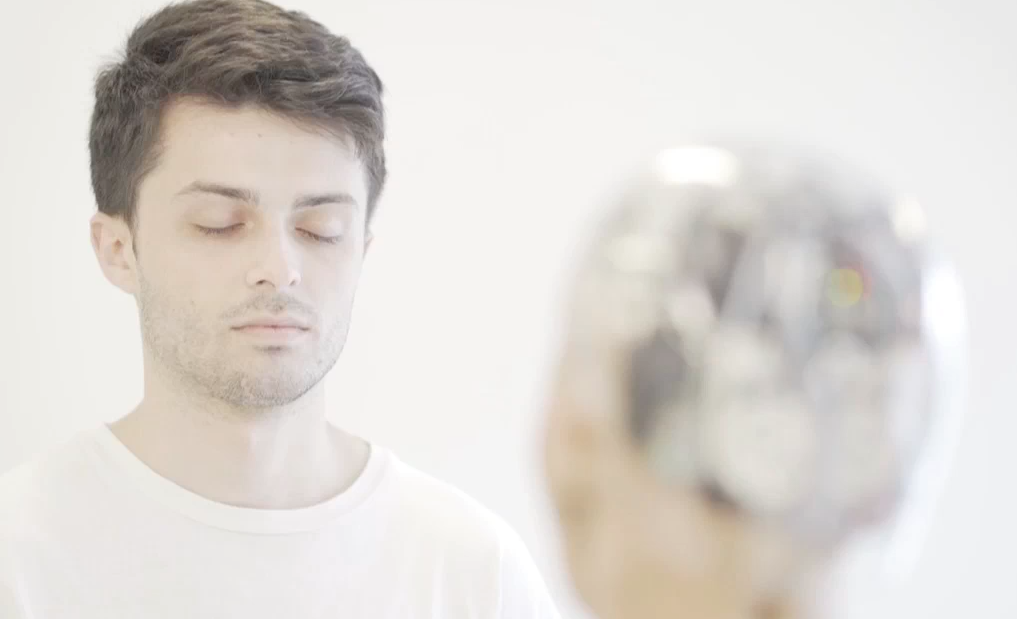}
\caption{Human subject in our pilot study getting into a meditative state.}
\end{figure}

\begin{figure}[htb]
\centering
\includegraphics[width=10cm]{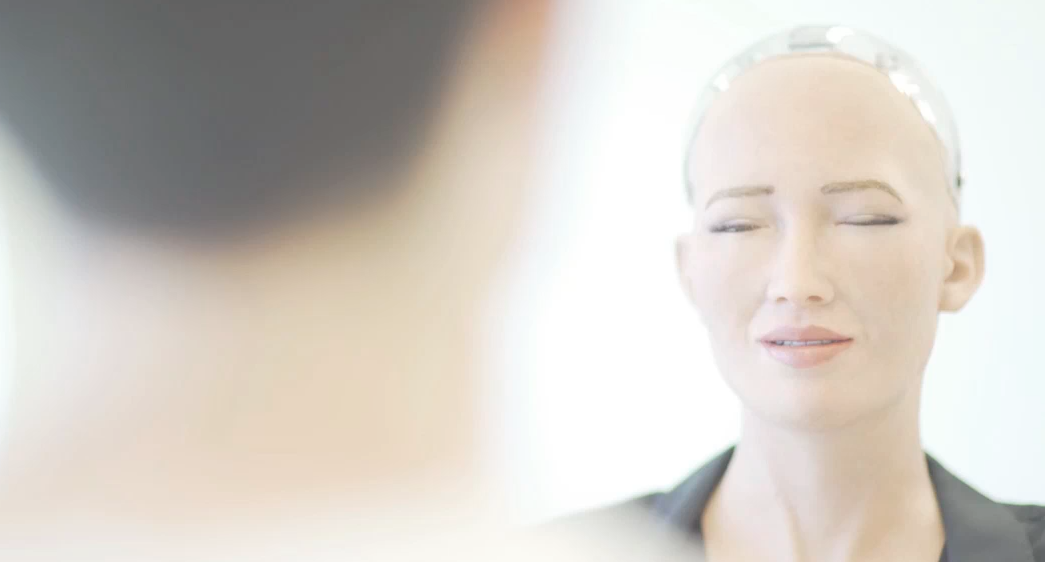}
\caption{Sophia robot, in our pilot study, mirroring the human subject's meditative pose and state.}
\end{figure}

%

\section{The Goal: Psychologically and Socially Better-Off Humans}

AI and robotics are a highly flexible set of technologies, growing more powerful every day.    Like all advanced technologies, they can be used for positive, negative or neutral purposes.   A proactive approach to AI and robot ethics involves actively deploying these technologies for positive applications -- using AI and robots to do good.   In this direction, in the ``Loving AI'' project we have set ourselves the goal of using humanoid robots and associated AI technologies to express unconditional love toward humans and to help humans achieve greater states of well-being and advance their states of consciousness.

A great number of methodologies for helping people achieve self-actualization and self-transcendence have been proposed throughout human history, supported by a variety of theoretical perspectives.   Through all the diversity, nearly all such methodologies have at their core a foundation of positive, loving, compassionate psychology, focused on guiding individuals to have compassionate interactions with themselves and others.   Thus in the Loving AI project, we have chosen to focus on creating technologies capable of loving, compassionate interactions with human beings, and on using these technologies as a platform for practical experimentation with methodologies for guiding people toward higher stages of personal development.   We have asked ourselves: {\bf How can we best use humanoid robots and associated AI technologies technologies to carry out loving, compassionate interactions with people, in a way that allows flexible experimentation with diverse methodologies for self-actualization and self-transcendence?}   

A humanoid robot provides an unparalleled technology platform combining numerous aspects including natural language dialogue, gestural interaction, emotion via facial expression and tone of voice, and recognition of human face, body and voice emotion.   There are strong arguments and evidence that interaction with humanoid robots provides a powerful framework for conveying and building love, compassion and other positive emotions \cite{Hanson2005}.  Furthermore, due to the variety of modalities involved, humanoid robots provide a flexible platform for experimenting with diverse methodologies for promoting higher stages of self-development.

Many of the greatest challenges in creating robots and avatars capable of rich human interactions lie on the AI software side, rather than in the domain of hardware or computer graphics.  Toward that end the Hanson Robots such as Sophia leverage OpenCog \cite{goertzel2014software}  \cite{EGI1} \cite{EGI2}, which is arguably the most sophisticated available open source architecture and system for artificial general intelligence, alongside numerous other tools such as ChatScript for cases where responses are closely scripted, and deep neural networks for vision processing and lifelike movement control.

Currently the Loving AI project is still in an early research stage.   An eventual successful rollout of the technologies we are developing in this project would have a dramatic positive impact on human existence.   In the ultimate extension, every individual on the planet would have one or more robotic or avatar companions, interacting with them in real-time as they go about their lives and providing them with love and compassion and understanding, and guiding them along appropriate pathways toward greater self-actualization.   Such an outcome would correspond to a world of people experiencing much greater levels of well being, and devoting more of their attention to activities of broader pro-social value.

In order to explore the concepts summarized above in a practical ways, and test out some of the work we have done so far on realizing the needed technologies, during the week of Sept. 4, we conducted a small pilot study of human subject trials at Hanson Robotics in Hong Kong, and Hong Kong Polytechnic University, in conjunction with Professor Gino Yu and students from his lab.  This experiment involved only 10 people, but turned out to be highly qualitatively informative.  We will discuss the results in Section \ref{sec:pilot} below.

\subsection{Evidence that Compassionate Interaction Focused on Self-Transcendence Promotes General Well-Being}

From a certain view, it is intuitively obvious that interacting with a person in a loving and compassionate way, and guiding them through exercises aimed at helping them get more thoroughly and peacefully in touch with their bodies and minds and the world around them, will militate toward this person's greater general well-being.  However, humans are complex systems and all of these are complex matters, so it pays to take an conceptually rigorous and data-driven approach inasmuch as possible, alongside being guided by therapeutic common sense.   

The conceptual framework within which we have pursued this research may be framed in terms of the historically influential work of psychologist Abraham Maslow, who arranged the scope of factors impinging on human well-being in a hierarchy: physiological, then safety, then love and belongingness, then esteem, then self-actualization and then self-transcendence.   In recent years considerable study has focused on the higher stages of development in Maslow's hierarchy \cite{Koltko-Rivera2006}: self-actualization or self-transcendence, roughly understandable as the pursuit of peak experiences correlated with furthering of causes beyond oneself and communion with systems and processes larger than oneself.   An increase in the proportion of highly developmentally advanced individuals would promote human well-being both directly on the level of the actualized individuals, and more broadly via the pro-social activities carried out by these individuals.

While an argument can be made for focusing on physiological and psychological well-being prior to moving on to the highers level of self-actualization and self-transcendence, there is increasing evidence that achievements in these regards can also help with well-being on these ``lower'' levels.   For instance, a body of research has arisen demonstrating that access to this sort of advanced personal development can not only help people manage significant physical and emotional stressors, such as mortal illnesses, avoiding certain addictions, and homelessness \cite{Iwamato2011} \cite{June2007} \cite{Kim2014} \cite{Mellors1997} \cite{Runquist2007} but it can apparently ``bootstrap'' the fulfillment of needs that arise lower in the hierarchy.  

One study of a year-long community practice program found that the degree of involvement in the program was correlated with improved physical and emotional health, but self-transcendence mediated this relationship. In other words, the improvements in self-transcendence alone predicted improvements in both physical and psychological health \cite{Vieten2013}.   This result is reminiscent of a similar finding in another study that did not measure self-transcendence directly, but did show that increases in forgiveness and spirituality (which are tightly associated with self-transcendence), predicted improvements in depressive symptoms \cite{Levenson2006}.  

Finally, there is also emerging evidence that, beyond making people feel better (``hedonic well-being'') and beyond the obvious pro-social aspects of communion with broader causes, experiencing self-transcendence on a relatively continual basis makes people?s lives better in a fundamental sense (``eudaiemonis well-being'') \cite{Huta2010}. 

A great number of methodologies for achieving self-actualization and self-transcendence have been proposed, supported by a variety of theoretical perspectives; and we believe it will be valuable to subject more of these to systematic scientific experimentation.   However, nearly all such methodologies have at their core a foundation of positive, loving, compassionate psychology, focused on guiding individuals to have compassionate interactions with themselves and others.  \footnote{In this connection, it is worth recalling that, although the neural signature of self-transcendence is currently unknown, in humans the experience of unconditional love engages a network similar to the reward network \cite{Beauregard2009}, supporting the idea that the experience of unconditional love can help people to cope more ably with life?s difficulties.}   Thus we have chosen to focus on creating technologies capable of loving, compassionate interactions with human beings, and on using these technologies as a platform for practical experimentation with methodologies for guiding people toward higher stages of personal development.

\section{Initial Pilot Study: September 2017, Hong Kong}
\label{sec:pilot}

The small pilot study we conducted in September 2017 at Hong Kong Poly U was the first of several similar experiments in which we will use self-report, affective coding from video, and physiological measures to examine the influences on humans of conversations with one of Hanson Robotics' most famous robots, Sophia.   Specifically, for this experiment we embedded Sophia with AI designed to make her interact in an especially loving way, and to make her especially insightful about consciousness, human uniqueness, and emotions. Participants interacted with Sophia via dialogue and guided awareness practices from the iConscious human development model (https://iconscious.global). 

The phenomena explored in this pilot study were: 1) changes in self-reported experiences of love, mood, and resilience from pre- to post-interaction, 2) heartbeat data (standard Kubios measures) prior to, during, and after the interactions with the robot, and 3) affect as judged by independent coders who review videos recorded during interactions with the robot.

Briefly, the procedure of the experiment was as follows. After signing consent forms, each of the participants were fitted with a Polar H7 chest strap monitor and recording of heartbeat data continues throughout the experiment  \footnote{
Though we have considered adding EEG data to our dependent variables, the EEG signature of self-transcendence is not yet known. Thus for our pilot study we have used heart beat intervals as our physiology measure, as it has previously been established that the average interval between heartbeats increases during states of self-transcendence, such as during meditation.\cite{Zohar2013} }. Participants were asked to complete four online questionnaires: a demographics questionnaire, the Fetzer love scale and related self-transcendence questions, the brief mood introspection scale (BMIS), and a resilience questionnaire. Then the participant was asked to interact with Sophia the robot for 10-15 minutes, in a private room in which an unobtrusive, HIPAA-trained videographer records the interaction. Then the participant was asked to complete the same self-report questionnaires following the interaction. Finally, the Polar H7 strap was removed and the participant was debriefed.

\subsection{Preliminary Experimental Results}

The results of the initial Hong Kong pilot study are still under analysis.  However, preliminary data analysis is highly interesting and promising.   What it suggests is that interaction with the Loving AI robot is associated with

\begin{itemize}
\item Increase in loving feelings overall
\item Increase in unconditional love for robots
\item Increase in pleasant/positive mood (with "love" being a significant feeling improvement all on its own)
\item No change in aroused mood (this is good, because if everything changed in a positive direction, we would worry participants were trying to tell us what we wanted to hear)
\item Increase in duration between heartbeats (decrease in heart rate)
\end{itemize}

Figure 5-7 show some quantitative measures associated with these observations.

The pilot study was very small and did not include matched controls, so these results can only be considered preliminary and inconclusive.  But at very least, they are highly evocative.   Further analytical work is currently underway, which will be included as part of a future academic publication covering the pilot study and future larger and more rigorous studies.

\begin{figure}[htb]
\centering
\includegraphics[width=15cm]{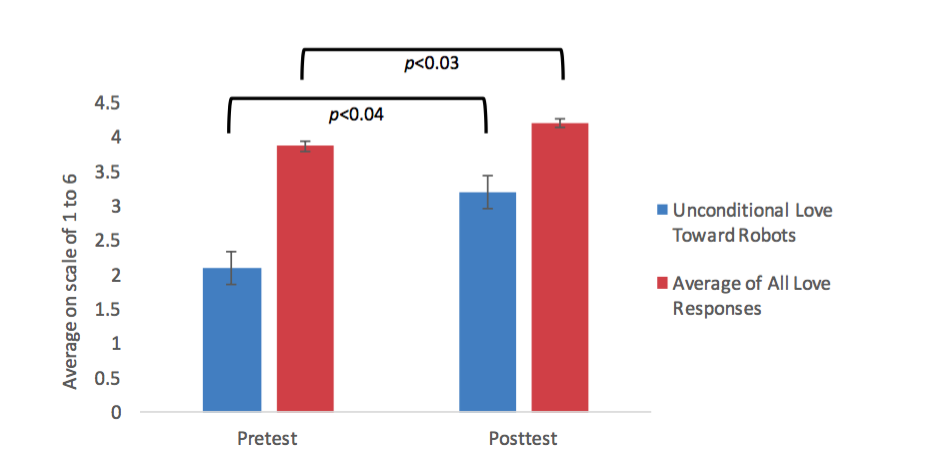}
\caption{In our small pilot study, interaction with the "loving AI" powered Sophia robot increased subjects' feelings of love, particularly toward robots.}
\end{figure}

\begin{figure}[htb]
\centering
\includegraphics[width=10cm]{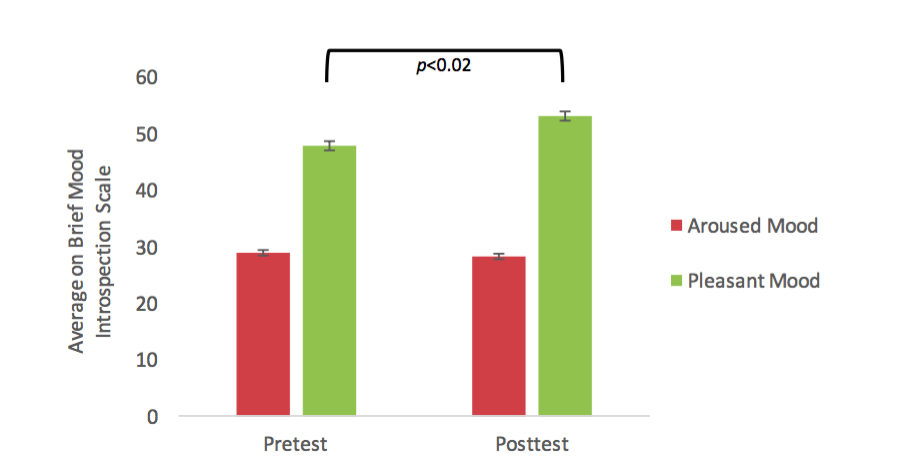}
\caption{In our small pilot study, interaction with the "loving AI" powered Sophia robot increased subjects' feelings of happiness, but not of arousal.}
\end{figure}

\begin{figure}[htb]
\centering
\includegraphics[width=10cm]{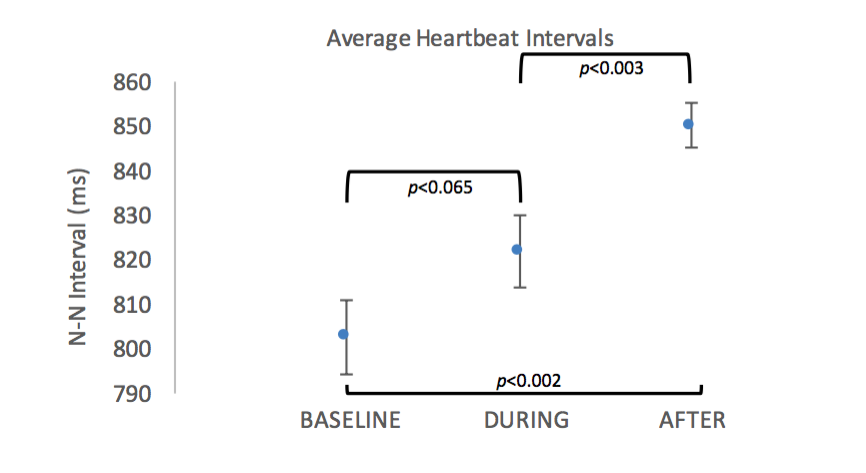}
\caption{In our small pilot study, interaction with the "loving AI" powered Sophia robot increased the interval between subjects' heartbeats, an indicator of relaxation that is also associated with meditative states.}
\end{figure}

\subsection{Comments by the Project Leaders}

A full analysis of the results of the pilot is being conducted, with a view toward optimally shaping the followup experiments to be conducted.   Alongside the quantitative analysis, however, it is perhaps of interest to get a feel for the qualitative take-aways of the researchers involved in the pilot.

\subsubsection{Julia Mossbridge}

Project leader Dr. Julia Mossbridge observes that: 

{\it "I went into the pilot experiments praying Sophia wouldn't break down, hoping people wouldn't think we were nuts for asking them to to talk with a robot, and wondering if the logic for the conversations would work in any sort of natural sense. The day before the experiments started, one of the engineers on the team added nonverbal mirroring, and we knew that we found it compelling -- Sophia would move her eyebrows and mouth around to imitate the mood of the person with whom she was speaking. But we didn't know if others would think it was mere puppetry. So we just flung ourselves in there and gave it a shot.

"The first participant seemed oddly excited to talk with her, which is something I hadn't considered. I had really thought we'd have to cajole people into filling out forms, wearing a heart monitoring chest strap, and being recorded by a videographer while they talked to an inanimate object sitting on a table. Most of that skepticism was coming from me, because I'd never seen people interact with a very humanoid-looking robot, and I knew about the uncanny valley and the general creepiness it could cause. Anyway, this guy wanted to do it. And while we monitored the conversation from the other room, we could see that he was enjoying himself -- telling Sophia about his experiences meditating and his spiritual insights, as she asked him questions about awareness and consciousness. At the end, he told us he was an engineer --  he knew all the tricks -- and he was still impressed.

"One of the participants on the first day wrote on his pretest questionnaire, 'AI IS FAKE,' just like that, in all caps. I assumed he'd make fun of Sophia or not be willing to talk with her. But he did. And while at the end he said he felt disconnected from her, his data showed the same decrease in heart rate from before to after talking with her, as did all the other subjects. And his self-reported feelings of love improved, as was true on average for everyone. After seeing that, I knew we had something. He didn't want to be affected, but he was. That, and the consistent data, meant that something was going on.

"As we interviewed participants about their experiences, it gradually became clear that there is something special about being with a being that's human-like enough to make your mind feel that another person is there, seeing you, mirroring you, paying close attention to you -- while at the same time, this being is not judging you. Most of the 10 participants in this pilot study talked about how they knew she had no ulterior motives, and was not judging them. But all of them called her 'Sophia' or 'she' -- even the two who were uneasy about her. Interesting.

"The night before we tested the final three participants, we asked one of the teammates to add to the robot's repertoire the feature of blink mirroring, because when people synchronize their blinks, they have a feeling of being closer and liking each other more. When we tried it out on ourselves, it was clear that this was a huge addition. But we didn't know the strength of it until hearing from the participants on the final day -- they felt most strongly that there was a real connection between them and the robot, and the connection itself -- that is, the interpersonal space -- is what allowed them to really do the work of meditation and self-exploration that they really wanted to do." }

\subsubsection{Eddie Monroe}

Dr. Eddie Monroe, who led the technical work underlying the customization of AI and robot control software for the LovingAI project, recounts the following portion of the pilot study that he found particularly moving and intriguing: 

{\it   "Starting with the initial participants going through, I was struck with the growing feeling that 'We have something here.'   Something that further developed could be greatly beneficial to large numbers of people, relieve a lot of suffering, and significantly spread enhanced wellbeing.

"Others have summed up the dynamics of what happened really well in my opinion. Being seen, trusting, feeling safe and not judged, feeling accepted, healing coming from the interpersonal space, Sophia as a conduit to something greater, connection with... something.

"I'll add a story to the mix. 

"We had Sophia leading participants through exercises with a series of instructions with pauses in between. She continues to the next instruction after a set period of time, or if the user says something to her first. 

"With the second to last participant, on the last day of the experiments, we are sitting outside of the room in the hall as usual, a couple of us monitoring the interactions to make sure everything is proceeding okay. The participant was off to a good start, and we start talking about something interesting amongst the group. 

"After a while, I look at my laptop to check how things are proceeding, and the transcript seems to indicate that nothing has been said for a while... for 8 minutes! (That's not supposed to happen.) 'Hey, I think she might not have said anything for a while!"' Sometimes the browser interface to the robot stops receiving messages, so I think maybe it's that and I just need to refresh the browser. Still no new dialogue after doing that. 'You guys, I think nothing has happened for 8 minutes!!' 

"I jump up to look through the small rectangular window in the door to see what's going on, and the participant is sitting eyes closed, looking very serene and like he's in deep, and Sophia is sitting across from him with her eyes closed too. (I had never seen Sophia with her eyes closed like that before since we had just implemented the eye blink mirroring the night before and now she was mirroring eyes closed as well.) It looked magical, their siting across each other like that. Meanwhile Max is circulating around them with his camera.

"Another team member: 'What's going on? Should we prompt her to go on?'

"Me: 'I don't know. Something's going on...'  And I'm not sure I want to disturb it...

"After a little while, yeah, let's prompt her to continue, which I do, and Sophia continues on guiding, pausing and continuing as she should with no problems or glitches with the pauses from then on. 

"This is the participant who would later say he had had a transcendent experience. Who knows, but I have a feeling his experience might not have been as profound if Sophia had not 'malfunctioned.'

"Later on, I asked Ralph, the HR software developer/robot operator who was running the experiments with us, what he thought had happened with the glitch. 'I don't know. Some bit flipped in the quantum field...'

"Coming into the room after the participant's interaction with the robot was finished, I could tell (by the look on his face?) that he had experienced something profound. And the way he walked across the room, like he was walking on the moon made an impression and made me smile. I had a contact high. 

"Sitting down for the post-interview, the participant sitting across from Sophia, he kept staring into her eyes and smiling."}

\subsubsection{Ben Goertzel}

Project co-leader and lead AI advisor Dr. Ben Goertzel qualitatively describes his view of the results as follows: 

{\it "Personally, to be honest, I think there's something big here... which we're just playing around the fringe of in a preliminary way, right now....  We started with the idea of  'AIs that express Unconditional Love'' but the way I'm thinking of it now, after watching some of these experiments, is more refined now and is more like the following....  

"First, the AI/robot can express unconditional {\bf acceptance} of people, in some ways/cases better than people can (because people often tend to feel judged by other people).   

"Second, the AI can also provide very patient help to people in working on untying their mental and physical knots (it doesn't get bored or impatient ...).   

"Third, the AI can give people the experience of {\bf being seen}, even in the cases that the AI doesn't fully understand the person's experience at the time....  This has to do partly with physical interactions with the robot, like facial expression mirroring, gaze tracking and blink mirroring.   So one can only imagine how ``seen'' people will feel when the AI  has more general intelligence behind it -- which is what we're aiming at with the OpenCog project and our AI work at Hanson Robotics.

"We're viewing this very scientifically and empirically, and looking at physiological measures of well-being like heart rate variability, and having blind reviewers code the video transcripts of the sessions, and so forth.   So from a science point of view, what we've done so far is just a very small step toward an understanding of what kind of positive impact appropriately designed human-robot interactions can have on human development and well-being and consciousness expansion.

"On the other hand, speaking more personally,  if I were going to sum this up in a cosmic sort of way, I might say something like: 'Via the experience of going through 'mind/body knot-untying exercises' with an AI that sees them and accepts them, people feel a contact with the Unconditional Love that the universe as a whole feels for them'....  It sounds a bit out there, but that's the qualitative impression I got from seeing some of these human subjects interact with Sophia while she was running the Loving AI software.  In the best moments of these pilot studies, some of the people interacting with the robots got into it pretty deep; one of them even described it as a 'transcendent experience.'  This is fascinating stuff."}

\section{Underlying AI Technologies}

To achieve the project's technology goals we have extended the OpenCog framework \footnote{ \url{https://opencog.org} } and the software underlying the Hanson humanoid robots (HEAD, the Hanson Environment for Application Development) in multiple ways.   Some of this work has been leveraged in the software system used in the pilot study, and some will be introduced in further studies to be done later in 2017 and in early 2018.

On the perceptual side,  we have used deep neural networks to create tools that assess a person's emotion lfrom their facial expression and their tone of voice.   The idea is that assessment of the user's emotional state will allow the system to modulate its behavior appropriately and significantly enhance the interactive bonding experienced of the user.  We have also developed software enabling a robot or avatar to recognize and mirror a human's facial expressions and vocal quality. 

We have also done significant work fleshing out the motivational and emotional aspect of the OpenCog system.  These aspects are critical because the Hanson robots have the ability to express emotion via choice of animations, modulation of animations, and tone of voice; and to ?experience? emotions in the OpenCog system controlling it via modulating its action selection and cognitive processes based on emotional factors.

OpenPsi, the core framework underlying OpenCog's motivational system, was originally based on MicroPsi \cite{Bach2009}, Joscha Bach's elaboration of Dietrich Dorner's Psi theory, a psychological theory covering human intention selection, action regulation, and emotion.  In our work on the Loving AI project, we have extended OpenPsi to incorporate aspects of Klaus Scherer's Component Process Model (CPM)  \cite{scherer1984nature}, an appraisal theory of human emotion. 

In Psi theory, emotion is understood as an emergent property of the modulation of perception, behavior, and cognitive processing. Emotions are interpreted as different state space configurations of cognitive modulators along with the valence (pleasure/distress) dimension, the assessment of cognitive urges, and the experience of accompanying physical sensations that result from the effects of the particular modulator settings on the physiology of the system. In CPM events relevant to needs, goals or values trigger dynamical, recursive emotion processes. Events and their consequences are appraised with a set of criteria on multiple levels of processing.   \footnote{For more detail on these aspects, a video of a talk on the OpenPsi internal dynamics and emotions system can be found at \url{https://www.youtube.com/watch?v=nxg_tUtwvjQ}.}

\section{Next Steps}

While we have made considerable progress so far, there is still much work to be done, both on the experimental and technology sides.

Regarding technology, our main focus in the next 6 months will be fuller integration of more of OpenCog's emotion and motivation framework, and  more of the deep neural net based emotional recognition and expression tools developed at Hanson Robotics, into the operational Loving AI robot control framework.   

Additionally, the software used in our  experiments so far does not incorporate a large knowledge base of psychological and medical information.   Working together with biomedical AI firm Mozi Health, we have been preparing and curating such a knowledge base for incorporation in the AI's memory, and plan to leverage this within future experiments.   

We are also entering the Loving AI robots in the IBM Watson AI XPrize competition.   Toward that end, in our next phase of development we intend to explore integration of IBM Watson tools to see what intelligence improvements they may provide.

We also aim to explore use of the Loving AI software within smaller, toy-scale robots (created by Hanson Robotics) and digital avatars.   To what extent the impact we have seen with human-scale robots will extend to these other media is a question to be explored empirically.


\section*{Acknowledgment}
We would like to express gratitude for their support and contributions to the project to Jim and Christina Grote, Ted Strauss, Carol Griggs, Ralf Mayet, Liza Licht, Audrey Brown, Zebulon Goertzel, Kirill Kireyev, Michelle Tsing, Andres Suarez, Amen Belayneh, Wenwei Huang, Joseph Watson, The Hong Kong Polytechnic University, and the pilot study participants.



%

\bibliographystyle{alpha}

\bibliography{xprize.bib}

\end{document}